\documentclass[review]{elsarticle}

\usepackage{lineno,hyperref}
\modulolinenumbers[5]

\journal{Pattern Recognition Letters}

\usepackage{geometry}                		
\usepackage{mathtools}
\usepackage{amsmath}
\usepackage{bm}
\usepackage{graphicx}
\usepackage{colortbl}
\usepackage{caption}
\usepackage{subcaption}
\usepackage{amssymb}
\usepackage[noend]{algorithmic} 
\usepackage[ruled,longend]{algorithm2e}
\usepackage{epstopdf}
\usepackage{color}
\definecolor{maroon}{cmyk}{0,0.87,0.68,0.32}

\newtheorem{lemma}{Lemma}[section]

\newtheorem{assumption}{Assumption}[section]

\newcommand{\bfX}{{\bf X}}
\newcommand{\bX}{{\bf X}}

\newcommand{\Var}{\mbox{Var}}

\newcommand{\bJ}{{\bf M}}

\newcommand{\bG}{{\bf G}}

\newcommand{\bfmu}{{\boldsymbol \mu}}

\newcommand\be{\begin{eqnarray}}
\newcommand\ee{\end{eqnarray}}
\newcommand\bee{\begin{eqnarray*}}
\newcommand\eee{\end{eqnarray*}}

\newcommand{\Ex}{\mathbb{E}}
\newcommand{\cov}{\textrm{cov}}

\newcommand{\var}{\textrm{var}}

\newcommand{\btheta}{\bm \theta}

\newcommand{\bj}{{\bf m}}
\newcommand{\bc}{\boldsymbol{\cdot}}

\usepackage{bbm}
\def\1{\mathbbm{1}}

\def\wh{\widehat}

\def\trans{^{\rm T}}
\def\th{^{\rm th}}
\def\bt{\begin{tcolorbox}}
\def\et{\end{tcolorbox}}
\usepackage{color, xcolor}
\definecolor{maroon}{rgb}{0.5,0,0.5}
\definecolor{green}{rgb}{0.3,0.5,0.1}
\definecolor{chestnut}{rgb}{0.8, 0.36, 0.36}
\definecolor{airforceblue}{rgb}{0.36, 0.54, 0.66}
\definecolor{aurometalsaurus}{rgb}{0.35, 0.50, 0.7}
\definecolor{bittersweet}{rgb}{1.0, 0.44, 0.37}
\definecolor{blush}{rgb}{0.87, 0.36, 0.51}
\definecolor{brickred}{rgb}{0.8, 0.25, 0.33}
\definecolor{bole}{rgb}{0.47, 0.27, 0.23}
\definecolor{auburn}{rgb}{0.43, 0.21, 0.1}
\definecolor{bleudefrance}{rgb}{0.19, 0.55, 0.91}

\newcommand{\bR}{\bf G}

\newcommand{\bdelta}{\bm \delta}

\newcommand{\bTheta}{\bm \Theta}

\newcommand{\bGamma}{\bm \Gamma}

\begin{document}

\begin{frontmatter} 

\title{Using the left Gram matrix to cluster high-dimensional data}

\author[mymainaddress, mysecondaryaddress]{Shahina Rahman} 

\author[mymainaddress, mysecondaryaddress]{Valen E. Johnson\corref{mycorrespondingauthor}}
\cortext[mycorrespondingauthor]{Corresponding author}
\ead{vejohnson@exchange.tamu.edu}

\author[mymainaddress, mysecondaryaddress]{Suhasini Subba Rao}

\address[mymainaddress]{Department of Statistics, Texas A \& M University}
\address[mysecondaryaddress]{3143 TAMU,
College Station, TX 77843-3143}


\begin{abstract}
For high dimensional data, where $P$ features for $N$ objects ($P \gg N$) are represented in an $N\times P$ matrix ${\bf X}$, we describe a clustering algorithm based on the normalized left Gram matrix, $ {\bf G} = {\bf  XX}^T/P$. Under certain regularity conditions, the rows in ${\bf G}$ that correspond to objects in the same cluster converge to the same mean vector.  By clustering on the row means, the algorithm does not require preprocessing by dimension reduction or feature selection techniques, and does not require specification of tuning or hyperparameter values. Because it is based on the $N\times N$ matrix {\bf G}, it has lower computational cost than many methods based on clustering the feature matrix ${\bf X}$. When compared to $14$ other clustering algorithms applied to  $32$ benchmarked microarray datasets, the proposed algorithm provided the most accurate estimate of the underlying cluster configuration more than twice as often as its closest competitors.  
\end{abstract}


\begin{keyword}
Clustering \sep Gram Matrix \sep High-Dimensional Features \sep Unsupervised
\end{keyword}


\end{frontmatter}

\linenumbers

\section{Introduction}

Despite their ubiquity in real applications, clustering of objects based on high dimensional features remains a challenging unsupervised learning task, made harder by the fact that the number of clusters is seldom known a priori. To detect clusters in high dimensional data, many clustering methods rely on preprocessing steps like feature selection or dimension reduction. Feature selection techniques are useful in finding clusters in sparse settings, where underlying clusters differ only by the values of a small number of features. However, the success of these techniques often depends on the astute selection of tuning parameters, which can be problematic in unsupervised settings. Dimension reduction techniques, like PCA, SVD and matrix factorization, perform well when underlying assumptions apply, but can fail to preserve cluster structure when they don't.  \citep{chang1983using}.

Our goal in this article is to describe an algorithm to cluster $N$ objects based on $P$ features when $P \gg N $. When $P$ features for $N$ objects are represented in an $N\times P$ matrix ${\bf X}$, we base the algorithm on the $N\times N$ left Gram matrix, ${\bf  XX}^T$.  Under certain regularity conditions, we show that this lower dimensional matrix preserves the cluster structure of the data. Standard clustering algorithms can then be applied to accurately identify the clusters at a reduced computational cost. The proposed algorithm does not require specification of tuning parameters or hyperparameters. 

\section{Notation and framework}\label{sec:notation}

We denote matrices by upper case bold letters (e.g., $\bX$) and column vectors by lower case bold letters (e.g., {\bf u}). We use $\1_P$ to denote the P-dimensional vector of ones and $\mathbbm{I}_A$ to denote the indicator function which equals $1$ if $A$ is true and $0$ otherwise. We let $||\bc||_2$ denote the Euclidean distance of a vector or Frobenius norm of a matrix, and $\|\cdot\|_{1}$ denote the absolute sum of the entries of a vector or matrix. We use $|\mathcal{S}|$ to denote the cardinality of the set $\mathcal{S}$. We write $i\simeq j$ if objects $i$ and $j$ are in the same cluster, and $i \not\simeq j$ otherwise.

Let $K_0$ be the (possibly unknown) number of true clusters. Let ${\bf f}_i = (x_{i,1},\ldots,x_{i,P})$ denote the $P$-dimensional feature vector measured on the $i\th$ object and stack the feature vectors of $N$ objects into an $N\times P$ feature matrix ${\bX}$. For a given $K_0$, define $\delta_i $ to be an integer in $\{1,\cdots, K_0\}$ that denotes the cluster membership of the $i\th$ object. We denote $\bdelta = (\delta_1,\ldots, \delta_N)$ as the vector of cluster identifiers, where $ Pr(\delta_i = a)= w_{a}$ with $\sum_{a=1}^{K_0} w_a = 1$. We assume that $\{{\bf f}_{i}\}_{i = 1}^N$ are conditionally independent $P$ dimensional random vectors with $\Ex({\bf f}_{i}|\delta_{i} = a) = {\bm \mu}_a$, where $\bfmu\trans_{a} = (\mu_{a,1},\ldots,\mu_{a,P})$ and finite covariance matrix $\Var({\bf f}_{i}|\delta_{i} = a) = {\boldsymbol \Sigma}_{a}$. We denote the vector of diagonal entries of ${\boldsymbol \Sigma}_{a}$ as ${\bm d}\trans_a = (\sigma^a_1,\ldots, \sigma^a_P)$, and we define $\theta_{a,b} =  {\bm\mu}_{a}\trans{\bm\mu}_{b}/P$ for $a\neq b$ and $\theta_{a} = {\bm\mu}_{a}\trans{\bm\mu}_{a}/P + {\bm d}\trans_a\1_{P}/P$, where $1 \leq a,b \leq K_0$.  




\section{The clustering algorithm}

The goal of our clustering algorithm is to partition $N$ objects into $K_0$ clusters using a transformed left Gram matrix, rather than directly from the ${\bf X}$ matrix. We assume that $K_0$ is unknown. The Gram matrix or similarity matrix in general is a central quantity for many clustering algorithms. However, as far as we are aware the following transformation has not been previously exploited.

\subsection{The left Gram matrix and it's properties}

    In the first step of the algorithm, we standardize each column of ${\bfX}$ to have mean 0 and standard deviation 1.  Using the standardized $\bfX$, we construct the matrix $\bR$ according to 
            \begin{equation}\label{eq:Rmat}
            \bR = {\bf X X}\trans/P.   
            \end{equation}

\noindent 
We denote the $(i,j)\th$ entry of the $\bR$ matrix by $g_{i,j}$.

If $\delta_i = a$ and $\delta_j = b$, $a,b \in {1,\ldots,K_0}$, then the expectation of the entries of the $\bR$ matrix is
\begin{eqnarray}
\label{eq:theta_ab}
\Ex(g_{i,j}|\delta_{i}=a,\delta_{j}=b) = \theta_{a,b}  \qquad
\textrm{ and } \qquad
\Ex(g_{i,i}|\delta_{i}=a) = \theta_{a}.
\end{eqnarray}
where, $ \theta_{a,b} = \sum_{p=1}^P\mu_{a,p}\mu_{b,p}/P$ and $\theta_{a} = \sum_{p=1}^P (\mu_{a,p}^2 + \sigma^{a}_{p})/P$.
This pivotal property of the $\bR$ matrix motivates us to detect the underlying clusters efficiently using the following two transformations.  

\subsection{The transformation on the left Gram matrix}
Next, we modify the elements of $\bG$ to form a matrix $\bJ$ by first appending the diagonal entries of $\bG$ as an additional column to $\bJ$.  In their place, we substitute the column-wise average from $\bG$.  The resulting $\bJ = \{ m_{i,j}$\} matrix is an $N\times (N+1)$ matrix with entries 

\begin{equation*}
    m_{i,j} = \begin{cases}
			g_{i,j}, & \text{for $j \neq i = 1, \ldots, N$ }\\
            g_{i,i}, & \text{for $j = N+1$}\\
            \frac{1}{N-1} \sum_{\substack{j = 1 \\j\neq i}}^N g_{i,j}, & \text{for $j = i$}.
		 \end{cases} 
\end{equation*}
We denote the $(N+1)$ dimensional $i\th$ row of $\bJ$ by ${\bf m}_i$.  

To illustrate the above transformations, we consider an example with $N=4$ objects where objects 1 and 2 belong to cluster 1, and objects 3 and 4 belong to cluster 2. 
The transformation from $\bG$ to $\bJ$ is 
\sloppy
\[ \label{eq:Jexample}
\bR = \begin{bmatrix}
    {\color{red} g_{1,1}} & g_{1,2} & g_{1,3}& g_{1,4} \\
    g_{2,1} & {\color{red} g_{2,2}} & g_{2,3}& g_{2,4} \\
    g_{3,1} & g_{3,2} & {\color{red} g_{3,3}}& g_{3,4} \\
    g_{4,1} & g_{4,2} & g_{4,3}& {\color{red} g_{4,4}} \\
    \end{bmatrix}
    \Rightarrow 
    \bJ = \begin{bmatrix}
            {\color{blue}\frac{g_{2,1}+g_{3,1}+g_{4,1}}{3} }  & g_{1,2}  & g_{1,3} & g_{1,4}& {\color{red} g_{1,1}} \\
            g_{2,1} & {\color{blue}\frac{g_{1,2}+g_{3,2}+g_{4,2}}{3} }  & g_{2,3} & g_{2,4}& {\color{red} g_{2,2}} \\
            g_{3,1} & g_{3,2} & {\color{blue}\frac{g_{1,3}+g_{2,3}+g_{4,3}}{3} } & g_{3,4} &  {\color{red} g_{3,3}} \\
            g_{4,1} & g_{4,2} & g_{4,3} &  {\color{blue}\frac{g_{1,4}+g_{2,4}+g_{3,4}}{3}} & {\color{red} g_{4,4}} \\
\end{bmatrix}
        \]
        
The transformation from $\bR$ to $\bJ$ matrix and equation \ref{eq:theta_ab} implies that the expected values of rows of $\bJ$ are equal for objects in the same cluster, with the exception of the diagonal elements. Hence ${\bf M}$ is useful in initializing our clustering algorithm when there is no prior information on the cluster identifiers, $\bdelta$.  

\subsection{ Updated modification of the Gram matrix }

At the model selection stage, we propose further modification of $\bJ$ matrix to $\bJ^{\wh\bdelta}$ based on the knowledge of $\wh\bdelta$ obtained from clustering the rows of $\bJ$ matrix. With this modification, expectation of all the elements of the $i\th$ and the $j\th$ row of $\bJ^{\wh\bdelta}$ become equal for all the columns whenever the corresponding $i\th$ and $j\th$ objects belong to the same cluster. For a given estimate of cluster identifiers $\widehat{\bdelta}$ from $\bJ$ matrix, we construct a matrix $\bJ^{\wh{\bdelta}}$ in which the diagonal elements of $\bJ^{\wh{\bdelta}}$ are updated with the column means restricted only to the rows of the remaining objects having the same cluster identifier , i.e., 
\begin{equation*}
    m_{i,i}^{{\widehat{\bdelta}}} = \sum_{\substack{j = 1 \\j\neq i}}^N g_{j,i}.\mathbbm{I}_{\{\wh{\delta}_j = \wh{\delta}_i\}}/\sum_{\substack{j = 1 \\j\neq i}}^N\mathbbm{I}_{\{\wh{\delta}_j = \wh{\delta}_i\}}.
\end{equation*}

For the above example in \ref{eq:Jexample} the transformation from $\bR$ matrix to $\bJ^{\bdelta}$ is
\begin{eqnarray*}
        \label{eq:Jdeltaexample}
            \bR = 
            \begin{bmatrix}
            {\color{red} g_{1,1}} & g_{1,2} & g_{1,3}& g_{1,4} \\
            g_{2,1} & {\color{red} g_{2,2}} & g_{2,3}& g_{2,4} \\
            g_{3,1} & g_{3,2} & {\color{red} g_{3,3}}& g_{3,4} \\
            g_{4,1} & g_{4,2} & g_{4,3}& {\color{red} g_{4,4}} \\
            \end{bmatrix}
            \Rightarrow 
            \bJ^{\bdelta} = 
            \begin{bmatrix}
            {\color{blue}g_{2,1}}  & g_{1,2}  & g_{1,3} & g_{1,4}& {\color{red} g_{1,1}} \\
            g_{2,1} & {\color{blue}g_{1,2} }  & g_{2,3} & g_{2,4}& {\color{red} g_{2,2}} \\
            g_{3,1} & g_{3,2} & {\color{blue} g_{4,3}} & g_{3,4} &  {\color{red} g_{3,3}} \\
            g_{4,1} & g_{4,2} & g_{4,3} &  {\color{blue} g_{3,4} } & {\color{red} g_{4,4}} \\
            \end{bmatrix}
            .
        \end{eqnarray*}
 
When $\wh\bdelta = \bdelta$, the expected values of the rows of ${\bf M}^{\bdelta}$ are described in the following lemma. 

\begin{lemma}\label{lemma:Jmean}
If $\delta_i=a$ (i.e., object $i$ belongs to cluster a), then the expectation of the $i\th$ row of $\bJ^{\bdelta}$ is the following $(N+1)$ dimensional vector
\begin{equation}\label{eq: Mmean}
\Ex\left({\bf m}^{\bdelta}_i| \delta_i = a\right) = \btheta_a \equiv  \Big(\{\theta_{a,\delta_j}\}_{j=1}^N \quad ,  \theta_{a}\Big) 
\end{equation}
\end{lemma}
Let $\bTheta^{N\times N+1} = \Ex\left[ {\bJ}^{\bdelta} |{\boldsymbol \delta}\right]$, where $\bTheta = 
     \left(\btheta_{\delta_1}, \cdots, \btheta_{\delta_N}\right)\trans
 $. Here $\bTheta$ represents the cluster means in the transformed space. The above transformation of $\bJ$ matrix makes the expected value of ${\bf m}^{\bdelta}_i$ and ${\bf m}^{\bdelta}_j$ equal whenever the $i\th$ and $j\th$ object belong to the same cluster. 
 This facilitates various model selection criterion, including , the Bayesian information criterion (BIC) \citep{schwarz1978estimating}, to select the number of clusters by minimizing the discrepancy between   $\bJ^{\delta}$ and the assumed model.

Next, we assume the following condition holds on the distinct cluster means on the transformed space.

\begin{assumption}\label{Ass:separability}
For $a \neq b \in \{1, \ldots, K_0\}$, there exists $\eta >0$ such that for all $P>0$
\begin{equation}\label{idcond}
    || \btheta_a - \btheta_b ||_2 > \eta.
\end{equation}
\end{assumption}
Such a condition is required to ensure that distinct clusters can be identified.  It requires that the proportion of informative feature vectors measured on objects does not converge to 0 as $P$ grows.


       


\subsection{Clustering strategy based on the rows of \texorpdfstring{$\bJ$}{M}}

To estimate the underlying number of clusters $K_{0}$ and the cluster indicator vector $\bdelta$ we maximize a quasi mixture likelihood for each possible value of $K= 1, \ldots, K_{{\text max}}$. Here, $K_{\text{max}}$ is an user-defined upper bound on $K_0$, which may equal $N$ if a prior bound is not known.  Because the elements of $\bJ$ and $\bJ^{\bdelta}$ represent an average of $P$ pairwise products, under certain regularity conditions the rows of $\bJ$ and $\bJ^{\bdelta}$ converge to a multivariate normal distribution.  We therefore maximize a quasi mixture likelihood function of the form  
$L_K(\bJ) =  \log \prod_{i = 1}^N\sum_{\substack{k = 1}}^K w_k \phi\Big(\bj_{i} ; \btheta_{k},{\bm \Gamma}_k\Big)$, where
\[ 
        \phi\Big(\bj_i;\btheta_k,\bGamma_k\Big) = \frac{1}{(2\pi)^{N/2}\hbox{det}(\bGamma_k)^{1/2}}\times \hbox{exp} \left\{ - \frac{1}{2}\Big(\bj_i - \btheta_k\Big)\trans{\bGamma_k}^{-1}\Big(\bj_i - \btheta_k\Big)\right\}. 
\] 
 Here the mixing weights $w_{k}$ satisfy $w_k \geq 0$ and $\sum_{k = 1}^K w_k = 1$. 
 Let $\wh L_{K}(\bJ) = L_{K}(\bJ; \wh w_K, \wh\btheta_{K}, \wh\bGamma_K)$ denote the maximized quasi log-likelihood for an assumed value of $K$. 
 The EM algorithm \citep{dempster1977maximum} is used to obtain maximum likelihood estimates for the parameters of the mixture model and the latent cluster identifiers $\wh\bdelta$ given $K$.

We estimate the number of clusters by maximizing 
the Bayesian Information Criterion (BIC) \citep{schwarz1978estimating}, which can be expressed as 
        \begin{eqnarray*}\label{eq:bic}
        \text{BIC}_{K} =  2\wh L_{K}(\bJ^{\widehat{\bdelta}}) - \nu_{K}\log N ,
        \end{eqnarray*}
where 
$\nu_{K}$ is the number of estimated parameters in the likelihood function $L_K(\cdot)$. 
\subsection{Implementation details}

The EM algorithm is only guaranteed to arrive at a local optimum of the mixture likelihood criterion \citep{mclachlan2004finite}. As a consequence, the choice of starting values for $\bdelta$ is important. Our experience and previous studies on multivariate mixture models by \citep{Mclust-software}, \citep{fraley2002model}, \citep{raftery2006variable} suggests that initial estimates of $\bdelta$ obtained from agglomerative hierarchical cluster analysis on the rows of $\bJ$ for a given K generally provide effective initialization. 
Pseudo code for implementing the clustering algorithm based on the modified Gram matrix, $\bJ$ and  $\bJ^{\bdelta}$ is described in Algorithm 1.   

\sloppy 

    \begin{algorithm}[ht!]
        \textbf{Input:}  $\bX \in {\mathbb R}^{N\times P}$ and $K_{\text{max}}$.\\ 
        \textbf{Output:} $\widehat{K}$; Cluster identifier, $\wh \bdelta = (\wh \delta_1, \ldots, \wh\delta_N)$. 
        \begin{enumerate}
        \item \textbf{G-step:} Construct the $N\times N$ similarity matrix, $\bR=\bX\bX^T/P$. 
        \item \textbf{M-step:} Rearrange the diagonal elements of $\bR$ to construct the $N\times (N+1)$ matrix $\bJ$. 
        \item {\bf for} $K = 1$ {\bf to} $K_{\text{max}}$: 
              \begin{enumerate}
                  \item [a.] \textbf{Initialize:} Use agglomerative hierarchical clustering on $\bJ$ to initialize the cluster identifier $\wh\bdelta $ having $K$ clusters. 
                  \item [b.] {\bf Repeat} 
                  \begin{enumerate}
                      \item \boxed{\bf Mstep:} {\bf for} $k = 1$ {\bf to} $K$ :
                      \item[] $n_k = \sum_{i =1}^N \mathbbm{I}_{\{\wh\delta_i = k\}}$; \hskip 4mm $\wh w_k = n_k/N$; \hskip 4mm
                      $\wh \btheta_k = \sum_{i = 1}^N \bj_i\cdot\mathbbm{I}_{\{\wh\delta_i = k\}}/n_k$; 
                      \item[] $\wh \bGamma_k = \sum_{i = 1}^N \left(\bj_i - \wh\btheta_k\right)\left(\bj_i - \wh\btheta_k\right)\trans\cdot\mathbbm{I}_{\{\wh\delta_i = k\}}/n_k$.
                      \item \boxed{\bf Estep:} {\bf for} $i = 1$ {\bf to} $N$: \hskip 2mm Set $\wh\delta_i = k$ \hskip 2mm for $  \arg\max_{k = 1}^K \wh w_k\phi\left(\bj_i; \wh\btheta_k, \wh\bGamma_k\right) = k$. 
                  \end{enumerate}
                  {\bf until} until $\wh\bdelta$ doesn't change. 
                  \item [c.] Given $\wh\bdelta$, calculate $\mbox{BIC}_K$ based on $\bJ^{\wh\bdelta}$. 
              \end{enumerate}
        \item {\bf Return} $K$ and $\wh \bdelta$ that maximize $\mbox{BIC}_K$ 
        \end{enumerate}
        \caption{GMCluster}
        \end{algorithm}

\section{Convergence of the \texorpdfstring{$\bJ^\delta$}{M} matrix}\label{sec:consistency}

    In the following lemma we obtain a rate of convergence for the mean squared error difference between $\bJ^{\wh\bdelta}$ and $\bTheta$. We require the following assumptions on the features. Recall that ${\boldsymbol \Sigma}_a = \Var({\bm x}_i|\delta_i = a)$, and let $\tau_{P} = \sup_{\substack{1\leq k \leq K_0\\}}\|{\boldsymbol \Sigma}_{k}\|_{1}^{1/2}$. For each feature $p = 1,\ldots,P$, let $y_{i,p} = x_{i,p}-\mu_{\delta_i, p}$ and define the $P$-dimensional vector ${\bf z}\trans_{i} = \left(y^2_{i,1},\ldots,y^2_{i,P}\right)$. Let ${\boldsymbol \Upsilon}_{a} = \Var({\bf z}_{i}|\delta_i = a)$ and $\kappa_{P} = \sup_{1\leq k\leq K_0}\|{\boldsymbol \Upsilon}_{k}\|_{1}^{1/2}$. 
            
            \begin{assumption}\label{assum:A}
            As $P\rightarrow \infty$,  we assume that $\kappa_{P}/P\rightarrow 0$ and $\tau_{P}/P\rightarrow 0$.  
            \end{assumption}
     We call the features weakly dependent if $\kappa_{P}$ and $\tau_{P}$ are $O(P^{1/2})$. This condition holds if the feature vectors are independent and have bounded variance. 
     


     \begin{lemma}\label{lemma:Jconsistency}
     Suppose Assumption 2 holds. Let $\mu_{\sup} =\sup_{a,p}|\mu_{a,p}|$ and $\sigma_{\sup} =\sup_{a,p}\sqrt{\sigma^{a}_{p}}$ and let $\mu_{\sup} , \sigma_{\sup} < \infty$. Then 
      \begin{equation*}\label{eq:Jconverge}
      \Ex\left\|\bJ^{\bdelta} - \boldsymbol{\Theta} \right\|_{2}^{2} \leq \Delta_{P}^{2} 
      \end{equation*}
      where $\Delta_{P} = \frac{1}{P}\left[N
      \{(N-1)\tau_{P}^{2}(2\mu_{\sup}+\sigma_{\sup})^{2} + (\kappa_{P}+2\tau_{P}\mu_{\sup})^{2}\}\right]^{1/2}$.
\end{lemma}
     
\noindent
The proof is in the supplementary material. 


This lemma suggests that for a correctly specified cluster configuration in which $\wh\bdelta = \bdelta$, $\bJ^{\bdelta}$ converges to $\bTheta$, the transformed cluster mean, in the order of $O(P^{-1/2})$, if the features are weakly dependent. For $K \leq K_0$, the identifiability condition (\ref{idcond})) guarantees that this sum-of-squares is bounded away from 0.  For $K>K_0$, the BIC penalty is sufficiently large to prevent sub-clusters from a given cluster from forming since the decrease in the sum-of-squares accumulated from such a split cannot offset a fixed penalty that is greater than $\log(N)$.  Thus, standard clustering algorithms are likely to be able to identify the correct cluster identifiers, provided that the conditions stated above are satisfied and $P$ is sufficiently large.  

\section{Application of the proposed method to bench-marked gene data sets}

Shah and Koltun \cite{shah2017robust} provided a recent comparison of several popular clustering methods for $32$ gene expression data sets based on adjusted mutual information (AMI) \citep{cover1991entropy}. The cluster configurations of these data sets are well studied, validated and are available in \href{https://schlieplab.org/Static/Supplements/CompCancer/datasets.htm}{\underline{DataLink}} \citep{de2008clustering}. 

To evaluate our method, GMcluster (GMC), we compared its performance to the clustering methods considered in \cite{shah2017robust}, adding $7$ additional state-of-the-art algorithms to the comparison.  Overall, we compared GMC with $12$ clustering algorithms that require pre-specification of the number of clusters and $6$ clustering algorithms that do not. As in \citep{shah2017robust}, for the methods that required pre-specification of the number of clusters we used those given in \citep{de9schliep}. Details of the methods considered and their parameter settings are provided in the supplementary section. 

\begin{figure}[ht!]
     \begin{subfigure}{0.55\textwidth}
         \includegraphics[scale = 0.43]{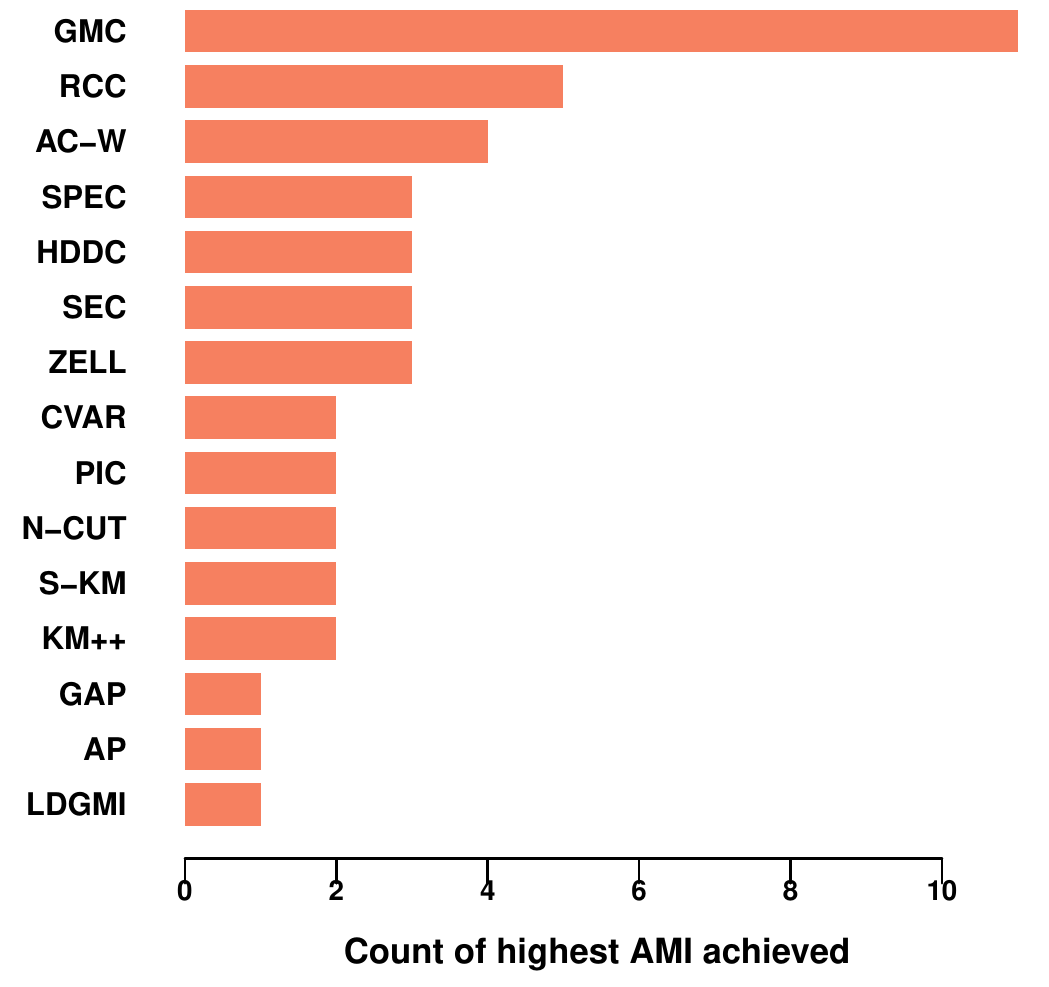}
         \caption{}
         \label{fig:AMI}
     \end{subfigure}
     \begin{subfigure}{0.23\textwidth}
         \includegraphics[scale = 0.26]{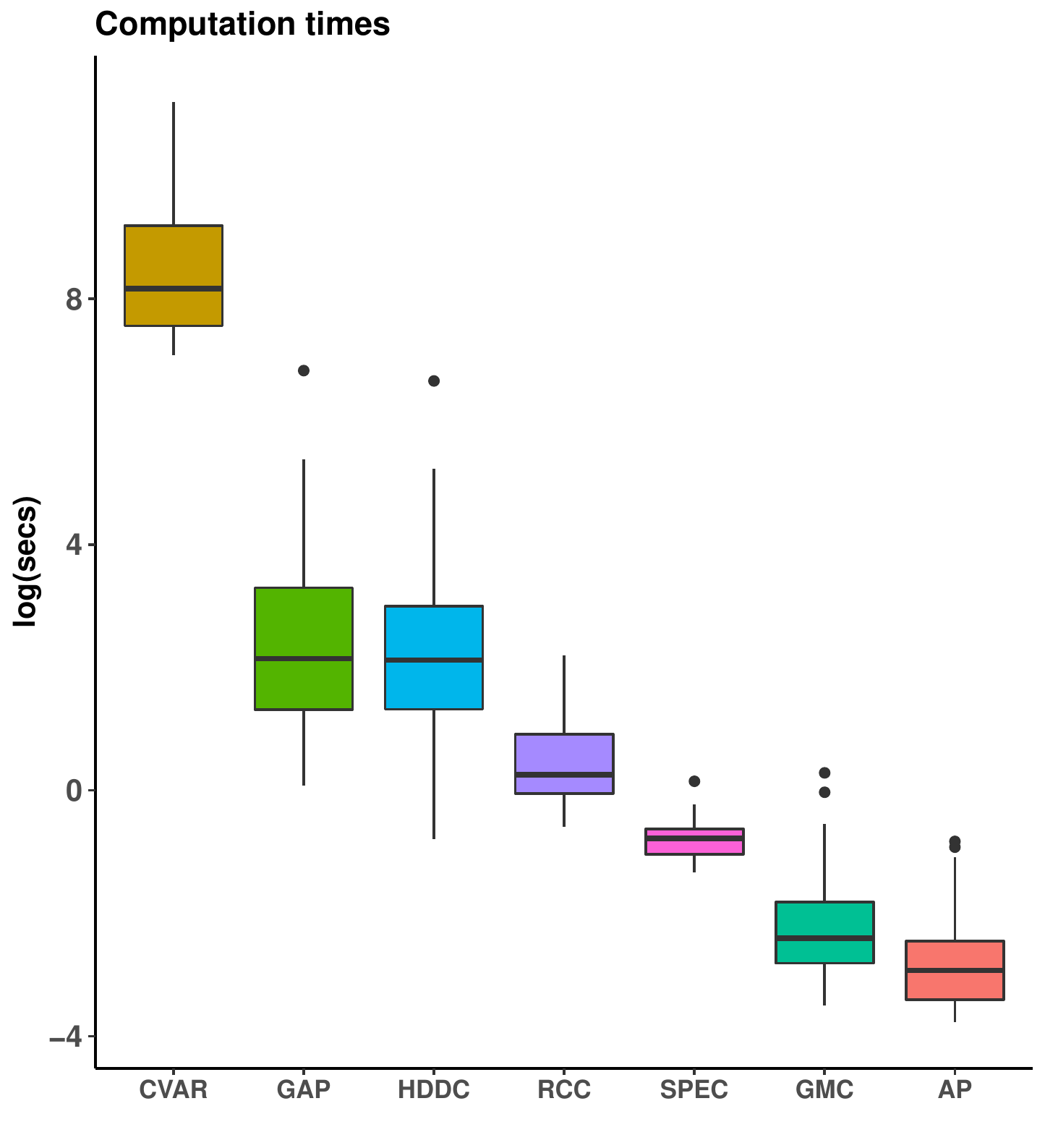}
         \caption{}
         \label{fig:Comp_times}
     \end{subfigure}
        \caption{(a) Barplot depicts the frequency of obtaining the highest accuracy (AMI) by each of the $14$ clustering algorithms across $32$ gene expression data sets. 
        (b) Boxplots display the distribution of computation times for the algorithms that do not require any predefined number of clusters ($K_0$).}
        \label{fig:Results}
\end{figure}

Figure \ref{fig:Results} summarizes the results of these comparisons. (For brevity, we have displayed the AMI results of only the $15$ algorithms that yielded the best AMI value for at least $1$ out of $32$ data sets in Table \ref{tableAMI} of the appendix.) The barplot depicts the number of times each method yielded the highest accuracy based on AMI. As the figure shows, GMC achieved the highest AMI for $11$ data sets, while the next best algorithm, RCC, provided the highest AMI in $5$ data sets. AMI values acheived by each clustering algorithm for each data set are provided in the Appendix. 

We also compared the computational performance of GMC to the six algorithms that, like GMC, also estimated the number of clusters. Because the runtime complexity of the GMC algorithm is linear in $P$, it was generally faster than the other algorithms (Figure \ref{fig:Results}b). Indeed, except for the AP algorithm (which only achieved the best AMI in one test data set), then GMC was faster by a factor of at least $7$ than all other algorithms that we tested. All comparisons were performed on a workstation with an Intel(R) Core(TM) i7-3770 CPU clocked at 3.40GHz with 8.00 GB RAM. The data sets and algorithms that produced these results are available at \href{https://github.com/srahman-24}{\underline{Github Link}}.

\section{Discussion}
Our clustering algorithm offers a simple and efficient method for clustering objects based on high-dimensional feature vectors. A software implementation of our proposed algorithm GMcluster is available in the R package, \href{https://cran.r-project.org/web/packages/RJcluster/index.html}{\it RJcluster} \citep{RJclust} \footnote{Our proposed algorithm has been implemented as RJcluster in CRAN as a R package}. 
The overall complexity of the current algorithm is $O(N^2P)$. Application to several genomic data sets suggest that the proposed algorithm provides a useful method for clustering objects when $N \ll P$. In ongoing  work, we are developing scalable techniques that will facilitate the use of our proposed algorithm in large $N$ and large $P$ settings. In future work, we will explore the utility of the proposed transformation on other similarity measures for meaningful clustering of unstructured data, like images or text documents.

\section*{Acknowledgments}
We thank Anirban Bhattacharya and Irina Gaynavova for their helpful comments and discussions, Marina Romanyuk for checking the computation times, and Rachael Shudde for maintaining the R package. Johnson and Rahman acknowledge support from NIH grant CA R01 158113. Subba Rao acknowledges the National Science Foundation grant DMS-1812054. 

\section*{References}
\bibliography{main_RJ.bbl}

\appendix

\section{Proof of Lemma \ref{lemma:Jconsistency}}
The proof of Lemma \ref{lemma:Jconsistency} follows from the proof of the following lemmas.

\begin{lemma}
{\bf Lemma A.1} Recall that $g_{i,j} = \sum_{p=1}^Px_{i,p}x_{j,p}/P$ with $\Ex(x_{i,p}|\delta_i) = \mu_{\delta_i,p}$ and $\Ex({\bf f}_{i}|\delta_i) = {\bm\mu}_{\delta_i}$ and $\Var({\bf f}_{i}|\delta_i) = {\bm\Sigma}_{\delta_i}$. Also recall that for clusters $a,b \in \{1, \ldots, K_0\}$, in equation \ref{eq:theta_ab} we defined  $\theta_{a,b} =  {\bm\mu}_{a}\trans{\bm\mu}_{b}/P$ and $\theta_{a} = {\bm\mu}_{a}\trans{\bm\mu}_{a}/P + {\bm d}\trans_a\1_{P}/P$, where ${\bm d}\trans_a = (\sigma^a_{1},\ldots, \sigma^{a}_{P})$ are the diagonal entries of ${\boldsymbol \Sigma}_{a}$. Then for $a,b \in \{1, \ldots, K_0\}$,
$\Ex[g_{i,j}|\delta_{i} = a,\delta_{j} = b] = \theta_{a,b} $ and 
$\Ex[g_{i,i}|\delta_{i}] = \theta_{a}$.
\end{lemma}

\noindent {\bf Proof:}

Suppose Assumption \ref{assum:A} holds. We define a $P$-dimensional vector ${\bf y}_{i} = (y_{i,1},\ldots,y_{i,P})$ with components $y_{i,p} = x_{i,p}-\mu_{\delta_i,p}$. It follows that $\Ex[y_{i,p}| \delta_i = a] = 0$ and $\Ex[y^2_{i,p}| \delta_i = a] = \sigma^{a}_{p}$. For $a \neq b \in \{1,\ldots, K_0\}$,  If $i\neq j$, $\delta_i = a$, and $\delta_j = b$, then
\begin{eqnarray*}
\label{eq:star}
g_{i,j} - \theta_{a,b}  &=&\frac{1}{P}\sum_{p=1}^{P}y_{i,p}y_{j,p}+
    \frac{1}{P}\sum_{p=1}^{P}\mu_{a,p}y_{j,p}
    +\frac{1}{P}\sum_{p=1}^{P}\mu_{b,p}y_{i,p} ,
\end{eqnarray*}
and for $i = j$
\begin{eqnarray*}
\label{eq:star2}
g_{i,i} - \theta_{a}
&=&\frac{1}{P}\sum_{p=1}^{P}(y_{i,p}^{2}-\sigma^{a}_{p})+ \frac{2}{P}\sum_{p=1}^{P}\mu_{a,p}y_{i,p}. \qquad
\end{eqnarray*}
Using the above and evaluating the conditional expectations $\Ex[g_{i,j}|\delta_{i},\delta_{j}]$ and 
$\Ex[g_{i,i}|\delta_{i}]$
proves
\begin{eqnarray}
\label{eq:mean}
\Ex(g_{i,j}|\delta_{i}=a,\delta_{j}=b) = \theta_{a,b} 
\textrm{ and }
\Ex(g_{i,i}|\delta_{i}=a) = \theta_{a}.
\end{eqnarray}

\begin{lemma}\label{lemma:cov}
{\bf Lemma A.2} Suppose Assumption 2 holds. Recall in section \ref{sec:consistency} for the $P$-dimensional vector ${\bf y}\trans_{i}$, we defined ${\boldsymbol \Upsilon}_{k} = \Var({\bf y}_{i}|\delta_i = k)$ and $\kappa_{P} = \sup_{1\leq k\leq K_0}\|{\boldsymbol \Upsilon}_{k}\|_{1}^{1/2}$ and $\tau_{P} = \sup_{\substack{1\leq k \leq K_0\\}}\|{\boldsymbol \Sigma}_{k}\|_{1}^{1/2}$. Let $\mu_{\sup} = \sup_{k,p}|\mu_{k,p}|$ and $\sigma_{\sup} = \sup_{k,p}\sqrt{\sigma^{k}_{p}}$. 
Then for $i\neq j$
\begin{eqnarray}
\label{eq:var}
\left(\Ex|g_{i,j} - \theta_{\delta_i,\delta_j}|^{2}\right)^{1/2} &\leq&
\frac{\tau_{P}}{P}(2\mu_{\sup}+\sigma_{\sup})
\end{eqnarray}
and 
\begin{eqnarray}
\label{eq:var2}
\left(\Ex|g_{i,i} - \theta_{\delta_i}|^{2}\right)^{1/2} &\leq&
\frac{1}{P}(\kappa_{P}+2\tau_{P}\mu_{\sup}).
\end{eqnarray}
Furthermore,
\begin{eqnarray}
\Ex\left\|{\bj}^{\bdelta}_{i} - \btheta_{\delta_i}\right\|_{2}^{2}
&\leq&
\frac{1}{P^{2}}
\left[(N-1)\tau_{P}^{2}(2\mu_{\sup}+\sigma_{\sup})^{2}
+ (\kappa_{P}+2\tau_{P}\mu_{\sup})^{2}\right],\label{eq:varseq}
\end{eqnarray} 
and 
\begin{eqnarray}
\label{eq:varmatrix}
\Ex\left\|\bJ^{\bdelta} - \boldsymbol{\Theta} \right\|_{2}^{2}
\leq \frac{N}{P^{2}}\left[(N-1)\tau_{P}^2(2\mu_{\sup}+\sigma_{\sup})^{2}+ (\kappa_{P}+2\tau_{P}\mu_{\sup})^{2}\right].
\end{eqnarray}
\end{lemma}


\noindent {\bf Proof:} 

We first prove (\ref{eq:var}), for the case $i\neq j$. 
We use (\ref{eq:star}) to give the bound
\begin{eqnarray*}
(\Ex|g_{i,j} - \theta_{\delta_i,\delta_j}|^{2})^{1/2} &\leq& 
\left[\Ex\left(\frac{1}{P}\sum_{p=1}^{P}y_{i,p}y_{j,p}\right)^{2}\right]^{1/2}
 + \left[\Ex \left(\frac{1}{P}\sum_{p=1}^{P}\mu_{\delta_i,p}y_{j,p}\right)^{2}\right]^{1/2} + \nonumber\\
&& \left[\Ex\left(\frac{1}{P}\sum_{p=1}^{P}\mu_{\delta_j,p}y_{i,p}\right)^{2}\right]^{1/2}.
\end{eqnarray*}
Recall that $\Ex(A^{2}) = \Ex[\Ex(A^{2}|\delta)]$ and if 
$\Ex[A|\delta]=0$, then $\Ex(A^{2}) = \Ex[\var(A|\delta)]$. This implies
\begin{eqnarray}
(\Ex|g_{i,j} - \theta_{\delta_i\delta_j}|^{2})^{1/2}
&\leq& A_{1,P}+A_{2,P}+A_{3,P},\label{eq:starbar}
\end{eqnarray}
where
\begin{eqnarray*}
A_{1,P}&=& 
\left(\Ex\left[\var\left(\frac{1}{P} \sum_{p=1}^{P}y_{i,p}y_{j,p}|\delta_{i},\delta_{j}\right)\right]\right)^{1/2}  \\
A_{2,P}&=& \left(\Ex\left[\var\left(\frac{1}{P}\sum_{p=1}^{P}\mu_{\delta_j,p}y_{i,p} |\delta_{i},\delta_{j}\right)\right]\right)^{1/2}  \nonumber\\
\textrm{and}\quad
A_{3,P}&=& \left(\Ex\left[\var\left(\frac{1}{P}\sum_{p=1}^{P}\mu_{\delta_i,p}y_{j,p} |\delta_{i},\delta_{j}\right)\right]\right)^{1/2}.
\end{eqnarray*}
We now bound each of the terms $A_{1,P},A_{2,P}$ and $A_{3,P}$. To bound $A_{1,P}$ we use the  
following decomposition:
\begin{eqnarray*}
&&\var\left(\frac{1}{P} \sum_{p=1}^{P}y_{i,p}y_{j,p}|\delta_{i}=a,\delta_{j}=b\right)\\
&=& \frac{1}{P^2}\sum_{p_1, p_2 = 1}^P \cov\left(y_{i,p_1}y_{j,p_1},y_{i,p_2}y_{j,p_2}| \delta_{i}=a,\delta_{j}=b\right) \\
&=& \frac{1}{P^{2}}\sum_{p_1,p_2=1}^{P}\cov[y_{i,p_1},y_{i,p_2}|\delta_{i}=a]\cov\left(y_{j,p_1},y_{j,p_2}|\delta_{j}=b\right) \\
&\leq& \sup_{a,p}\sigma_{a,p}\frac{1}{P^{2}}\sup_{a}\sum_{p_1,p_2=1}^{P}|\cov\left(y_{i,p_1},y_{i,p_2}|\delta_{i}=a\right)|
\leq \frac{\tau_{P}^{2}}{P^{2}}\sigma_{\sup}^{2}.
\end{eqnarray*}
It follows that
\begin{eqnarray*}
A_{1,P}\leq \left(\Ex\left[\var\left(\frac{1}{P} \sum_{p=1}^{P}y_{i,p}y_{j,p}|\delta_{i},\delta_{j}\right)\right]\right)^{1/2}
\leq \frac{\tau_{P}}{P}\sigma_{\sup}.
\end{eqnarray*}
Using a similar argument to bound the conditional variance inside $A_{2,P}$, we have
\begin{eqnarray*}
\var\left(\frac{1}{P} \sum_{p=1}^{P}\mu_{i,p}y_{j,p}|\delta_{i}=a,\delta_{j}=b\right) &\leq& \mu_{\sup}^{2}\frac{1}{P^{2}}\sum_{p_{1},p_{2}=1}^{P}
|\cov(y_{i,p_1},y_{i,p_2})|\leq \frac{1}{P^{2}}\mu_{\sup}^{2}\tau_{P}^{2}. \label{eq:Pall2}
\end{eqnarray*} 
This leads to 
\begin{eqnarray*}
A_{2,P} \leq \frac{\tau_{P}}{P}\mu_{\sup},
\end{eqnarray*}
and by a similar argument to
$A_{3,P} \leq \frac{\tau_{P}}{P}\mu_{\sup}$.
Substituting these bounds into (\ref{eq:starbar}) we obtain
\begin{eqnarray*}
\label{eq:rmnbound}
(\Ex|g_{i,j} - \theta_{\delta_i,\delta_j}|^{2})^{1/2}
\leq  \frac{\tau_{P}}{P}(2\mu_{\sup}+\sigma_{\sup}),
\end{eqnarray*}
thus proving (\ref{eq:var}).
We next bound $(\Ex|g_{i,i} - \theta_{\delta_i}|^{2})^{1/2}$. We use (\ref{eq:star2}) to give
\begin{eqnarray}
(\Ex|g_{i,i} - \theta_{\delta_i}|^{2})^{1/2} &\leq& 
 B_{1,P}+B_{2,P}+B_{3,P},\label{eq:starbar1}
\end{eqnarray}
where
\begin{eqnarray*}
B_{1,P}&=& 
\left(\Ex\left[\var\left(\frac{1}{P} \sum_{p=1}^{P}y_{i,p}^{2}|\delta_{i}\right)\right]\right)^{1/2}  \\
B_{2,P}&=& \left(\Ex\left[\var\left(\frac{1}{P}\sum_{p=1}^{P}\mu_{\delta_i,p}y_{i,p} |\delta_{i}\right)\right]\right)^{1/2}  \nonumber\\
\textrm{and}\quad
B_{3,P}&=& \left(\Ex\left[\var\left(\frac{1}{P}\sum_{p=1}^{P}\mu_{\delta_i,p}y_{i,p} |\delta_{i}\right)\right]\right)^{1/2}.
\end{eqnarray*}
Using the same methods used to bound $A_{2,P}$ and $A_{3,P}$, it is straightforward to show that $B_{2,P},B_{3,P}\leq \tau_{P}\mu_{\sup}/P$. To bound $B_{1,P}$ we note that 
\begin{eqnarray*}
\var\left(\frac{1}{P} 
\sum_{p=1}^{P}y^2_{i,p}|\delta_{i}=a\right) 
&=& \frac{1}{P^2}\sum_{p_1, p_2 = 1}^P \cov[y_{i,p_1}^{2},y_{i,p_2}^{2}|\delta_{i}=a] 
\leq P^{-2}\kappa_{P}^{2},
\end{eqnarray*}
which follows from Assumption \ref{assum:A}. Thus 
$B_{1,P} \leq P^{-1}\kappa_{P}$.
Substituting into (\ref{eq:starbar1}) gives 
\begin{eqnarray*}
\left(\Ex|g_{i,i} - \theta_{\delta_i})^{2}|\right)^{1/2} &\leq&
\frac{1}{P}(\kappa_{P}+2\tau_{P}\mu_{\sup}),
\end{eqnarray*}
thus proving (\ref{eq:var2}).

To prove (\ref{eq:varseq}), we apply (\ref{eq:var}) and (\ref{eq:var2}), leading to 
\begin{eqnarray*}
\Ex\left\|{\bf m}^{\bdelta}_{i} - \btheta_{\delta_i}\right\|_{2}^{2}
&=&
\sum_{i,j=1,i\neq j}^{N}\Ex(g_{i,j} - \theta_{\delta_i,\delta_{j}})^{2}
+\sum_{i=1}^{N}\Ex(g_{i,i} - \theta_{\delta_i})^{2}
\nonumber\\
&\leq& (N-1)\frac{1}{P^{2}}\tau_{P}^{2}(2\mu_{\sup}+\sigma_{\sup})^{2}+ \frac{1}{P^{2}}(\kappa_{P}+2\tau_{P}\mu_{\sup})^{2}
\nonumber\\
&\leq& \frac{1}{P^{2}}\left[(N-1)\tau_{P}^{2}(2\mu_{\sup}+\sigma_{\sup})^{2} + (\kappa_{P}+2\tau_{P}\mu_{\sup})^{2}
\right],
\label{eq:dragon}
\end{eqnarray*} 
proving (\ref{eq:varseq}).
Finally, to prove (\ref{eq:varmatrix}) we note that
\begin{eqnarray*}
\Ex\left\|\bJ^{\bdelta} - \boldsymbol{\Theta} \right\|_{2}^{2} = \sum_{i=1}^{N} \Ex\left\|{\bf m}^{\bdelta}_{i} - \btheta_{\delta_i}\right\|_{2}^{2}.
\end{eqnarray*}
By substituting (\ref{eq:varseq}) above we have
\begin{eqnarray*}
\Ex\left\|\bJ^{\bdelta} - \boldsymbol{\Theta} \right\|_{2}^{2}\leq
\frac{N}{P^{2}}
\left[(N-1)\tau_{P}^{2}(2\mu_{\sup}+\sigma_{\sup})^{2}
+ (\kappa_{P}+2\tau_{P}\mu_{\sup})^{2}\right],
\end{eqnarray*}
which proves (\ref{eq:varmatrix}) and hence Lemma \ref{lemma:Jconsistency}. 
\hfill $\Box$

\section{Description of other clustering algorithms}

We divided the competing methods into two categories. We consider $12$ algorithms that $require$ the pre-specification of the number of clusters $K_0$ : k-means++ ({\bf km++}) \citep{arthur2007k}, Gaussian mixture models ({\bf GMM}), fuzzy clustering ({\bf fuzzy}), mean-shift clustering ({\bf MS}) \citep{comaniciu2002mean},  agglomerative hierarchical clustering with ward linkage ({\bf AC-W}), normalized cuts ({\bf N-Cuts}) \citep{shi2000normalized}, Zeta l-links ({\bf Zell}) \citep{zhao2009cyclizing}, spectral embedded clustering ({\bf SEC}) \citep{nie2009spectral}, clustering using local discriminant models and global integration ({\bf LDMGI}) \citep{yang2010image}, path integral clustering ({\bf PIC}) \citep{zhang2013agglomerative}, sparse k-means ({\bf sp-km}) \citep{witten2010framework} and sparse subspace clustering ({\bf SSC}) \citep{elhamifar2013sparse}. We consider $6$ other clustering methods that {\it do not require} the pre-specification of $K_0$ : affinity propagation ({\bf AP}) \citep{frey2007clustering}, a robust graph continuous clustering ({\bf RCC}) \citep{shah2017robust}, GAP statistics with partitioning around medoids ({\bf GAP}) (\citep{tibshirani2001estimating}, \citep{kaufman1990partitioning}), a model based clustering with variable selection ({\bf Cvarsl}) \citep{raftery2006variable}, a high dimensional data clustering ({\bf HDDC}) \citep{bouveyron2007high} and a graph clustering based on tensors ({\bf SPEC}) \citep{Spectrum}. The hyperparameter settings used for these methods are provided in Table \ref{tab:hyper}. 

Figure 1 provides a summary of the number of times each algorithm achieved the highest AMI. This figure does not display results for  {\bf GMM}, {\bf MS}, {\bf fuzzy}, and {\bf SSC} algorithms because none of these algorithms provided the highest AMI for any data set.

\bigskip

{\bf Computation times}

\bigskip

We compared the execution time of our proposed method, GMcluster algorithm to other algorithms which {\it do not require} the prespecification of the number of clusters. The overall distribution of the computation times taken by these algorithms across $32$ datasets is displayed in Figure \ref{fig:Results}b. Execution times are displayed in log seconds. The {\bf AP}, {\bf HDDC}, {\bf RCC} and {\bf SPEC} algorithms are much more computationally efficient than {\bf GAP} and {\bf Cvarsl} methods.  Excluding the AP algorithm, which provided the best AMI in only one case, the GMcluster algorithm was atleast 7 times faster than all of the remaining algorithms. 

\bigskip

{\bf Hyperparameter settings of other clustering methods}

\bigskip

We used the same hyperparameters settings recommended in \citep{shah2017robust} for the following clustering algorithms:  {\bf KM++}, {\bf AC-W}, {\bf N-CUT}, {\bf ZELL}, {\bf SEC}, {\bf LDMGI}, {\bf PIC} and {\bf RCC} that.  Table \ref{tab:hyper} provides the hyperparameter settings and software used to obtain the results from the other methods. 

\begin{table}[!htbp] 
\caption{Hyperparameters and software used for other methods in the comparative study.} \label{tab:hyper}
\begin{center}
\resizebox{\columnwidth}{!}{%
\begin{tabular}{|c|c|c|c|}
     \hline 
    {\bf Methods}  &{\bf hyper-parameters}   & {\bf Values}  & {\bf Software} \\ \hline \hline 
    {\bf AP}      &iter.max     &100       & apcluster \\
                  &s           &negDistMat(r=2) &(R package) \\ \hline 
    {\bf CVAR}    &search      &headlong  &clustvarsel \\ 
                  &direction   &forward   &(R package)\\
                  &parallel    &T         & \\
                  &iter.max     &100       & \\ \hline 
    {\bf GAP}     & maximum number of clusters & 20 & cluster \\ 
    (with partitioning around mediod) &d.power &2&(R package)\\                            
                  &bootstrap samples &max(100,n)  &    \\
                  &metric    &Euclidean &cluster  \\ 
                  & iter.max   & 100      &         \\\hline 
    {\bf HDDC}    &max number of clusters  &20   &HDclassif \\
                  &model       &``ALL"      &(R package)  \\
                  &threshold   & 0.2(default)  &      \\ 
                  &criterion   & bic(default)  &      \\ 
                  &$d_{\text{max}}$   & 100(default)  &      \\\hline                   
    {\bf S-KM}    &iter.max  &100  &sparcl\\
                  &wbounds grid & [1, 10](default) &(R package)\\ 
                  &nperm        &100      &  \\\hline
    {\bf SPEC}    &method     &2(default: multimodal eigen gap) &Spectrum \\ 
                  &kernel-type &density(default) &(R package)\\
                  &maxk       &20       &           \\ 
                  &Nearest-Neighbor &7(default) &  \\
                  &iter.max    &100      & \\\hline  
    {\bf GMC}     &Cmax       &20      & RJcluster\\
                 &iter.max   &100     & \\ \hline \hline 
\end{tabular}
}
\end{center}
\end{table}

\bigskip

{\bf Data transformations} 

\bigskip

We took the logarithmic transformation of all data that contained only positive values. Several data sets were already preprocessed and centered and were therefore not log-transformed further. These data sets included Alizadeh-v1,v2,v3, Bittner, Garber, Lapointe-v1, Liang, Risinger, Singh-v1, Tomlins-v1 and West. For the remaining data sets, after logarithm transformation we standardized by centering on the median and scaling by the standard deviation. Further details regarding the transformations applied to each data set can be found in the ``scaling.R" folder in the RJclust folder provided in the github repository \href{https://github.com/srahman-24/GMclust}{https://github.com/srahman-24/GMclust.}

\bigskip

{\bf Data Availability and Software}

\bigskip

The datasets used for the comparisons are available at \href{https://schlieplab.org/Static/Supplements/CompCancer/datasets.htm}{\underline{DataLink}}. 
We executed all algorithms on a workstation with an Intel(R) Core(TM) i7-3770 CPU clocked at 3.40GHz with 8.00 GB RAM. The datasets and algorithms that produced the results are available at \href{https://github.com/srahman-24}{\underline{https://github.com/srahman-24}}.

\bigskip
{\bf AMI calculations}
\bigskip

\begin{table*}[h!]
\caption{Adjusted Mutual Information for $15$ clustering algorithms over $32$ gene expression datasets. For each dataset, the maximum achieved AMI is highlighted in bold. } \label{tableAMI}
\resizebox{\textwidth}{!}{%
\begin{tabular}{ |c c c | c c c c c c c c| c c c c c c c| } 
\hline
\hline
& & &\multicolumn{8}{c|}{$K_0$ is {\bf known}}&\multicolumn{7}{|c|}{$K_0$ is {\bf unknown}}\\
\hline
{\bf Datasets} &{\bf N} &{\bf P} 
&{\bf km++} &{\bf sp-km}  &{\bf AC-W}  &{\bf N-Cuts} &{\bf Zell} &{\bf SEC} &{\bf LDGMI}  &{\bf PIC} &{\bf AP} &{\bf GAP} &{\bf HDDC} &{\bf SPEC} &{\bf RCC} &{\bf Cvarsl} &{\bf GMC}\\ \hline \hline
{\bf Alizadeh-v1 }  &42  &1097 
&0.340	&-0.015	&0.101	&0.096 &0.250  &0.238  &0.123  &0.033 &0.211 &0.000 &0.133 &0.157  &0.426 &-0.006 &{\bf 0.515}\\ \hline
{\bf Alizadeh-v2 } &62  &2095
&0.568	&0.872 &0.922	&0.922	&0.922 &0.922  &0.738	&0.922 &0.563 &{\bf 1.000} &0.571	&0.753	&{\bf 1.000}
&0.533 &{\bf 1.000}   \\ \hline
{\bf Alizadeh-v3}  &62  &2095
&0.586	&0.689	&0.616	&0.601	&0.702	&0.574  &0.582  &0.625  &0.540 &0.678 &0.548 &0.609 &{\bf 0.792}
&0.295  &{\bf 0.792} \\ \hline
{\bf Armstrong-v1} &72   &1083
&0.372	&0.370	&0.308	&0.372  &0.308  &0.323  &0.355  &0.308  &0.381 &0.475 &0.461  &0.617 &0.546 &0.302 &{\bf 0.637}  \\ \hline
{\bf Armstrong-v2} &72    &2196
&{\bf 0.891}	&0.375	 &0.746	&0.83  &0.802 &{\bf 0.891} &0.509 &0.802 &0.586 &0.525 
&0.000 &0.693 &0.838  &0.513 &0.661 \\ \hline
{\bf Bhattacharjee} &203  &1545    
&0.444	&0.296	&{\bf 0.601}	&0.563	&0.496	&0.570  &0.378	&0.378 &0.377	&0.518 &0.000   &0.505	&0.600 &0.173  &0.453\\  \hline 
{\bf Bittner }   &38    &2203
&-0.012	&0.195	&0.002	&0.042	&0.115	&-0.002	&0.014	&0.115 &0.243 &0.000 & 0.288  &0.013	&0.156  &-0.020 &{\bf 0.341}\\  \hline 
{\bf Bredel} &28    &1072
&0.297	&0.000	&0.384	&0.203	&0.278	&0.259	&0.295	&0.278	&0.139 &0.035 &0.227 &0.356	&{\bf 0.466}  &-0.002 &0.265  \\ \hline
{\bf Chowdary}   &104   &184
&0.764	&0.595	&{\bf 0.859}	&{\bf 0.859}		&{\bf 0.859}	&{\bf 0.859}	&{\bf 0.859}	&{\bf 0.859} &0.443 &0.000 &0.625 &0.575	&0.393   &0.000 &0.585 \\ \hline
{\bf Dyrskjot}  &40    &1205
&0.507	&{\bf 0.755} &0.474	&0.303	&0.269	&0.389	&0.385  &0.177	&0.558 &0.348 &0.607 &0.629	&0.383  &0.292 & 0.742 \\ \hline
{\bf Garber}   &66    &4555
&0.242	&0.026	&0.210	&0.204	&0.246	&0.200	&0.191	&0.246 &{\bf 0.274} &0.096 &0.164 &0.137 &0.173 &0.175 &0.130 \\ \hline
{\bf Golub-v1} &72    &1870
&0.688 	&0.701	 &{\bf 0.831}	&0.650 	&0.615	&0.615	&0.615  &0.615 &0.430 &0.044  &0.478	&0.137	&0.490 &0.628  &0.420  \\  \hline
{\bf Golub-v2} &72    &1870
&0.680	&0.617	&{\bf 0.737}	&0.693	&0.689	&0.703	&0.600	&0.689 &0.516	&0.000 &0.478   &0.352	&0.597  &0.139 &0.538  \\  \hline
{\bf Gordon}   &181    &1628
&0.651	&{\bf 0.937}		&0.483	&0.681	&-0.005	&0.791	&0.669	&0.664	&0.304 &0.435
&0.000   &{\bf 0.937}	&0.343 &0.140 &0.499\\  \hline
{\bf Laiho}   &37    &2204
&0.007	&0.062	&-0.007	&0.030	&0.073	&-0.007	&0.093	&0.044	&0.061 &0.000 &{\bf 0.220}  &0.036	&0.000   &0.091 & 0.185\\ \hline
{\bf Lapointe-v1}  &69    &1627
&0.088	 &0.012  &0.151	&0.179	&0.151	&0.088	&0.149	&0.151	&0.162 &0.034&0.165  &0.012	&0.156 &{\bf 0.180 } &{\bf 0.181} \\  \hline
{\bf Lapointe-v2}  &110    &2498
&0.008	&0.097	&0.033	&0.153	&0.147	&0.028	&0.118	&0.171	&0.210 &0.199 &0.000  &-0.006	&{\bf 0.239}  &0.133 & 0.172 \\  \hline
{\bf Liang}  &37    &1413
&0.301	&0.301	&0.301	&0.301	&0.301 &0.301	&0.301	&0.301 &0.481 &0.243	&{\bf 0.523}   &0.301	&0.419  &0.296 &0.481\\ \hline
{\bf Nutt-v1}   & 50   &1379
&0.171	&0.311	&0.159	&0.156	&0.109	&0.086	&0.078  &0.113	&0.116 &0.000 &0.443 &0.215	&0.129  &0.112 &{\bf 0.459} \\  \hline
{\bf Nutt-v2}  &28    &1072
&-0.025	&0.000	&-0.024	&-0.025	&-0.031	&-0.025	&-0.027	&-0.030	&-0.027 &0.035 &0.152 &{\bf 0.250}	&-0.029 &-0.002 &{\bf 0.250} \\  \hline
{\bf Nutt-v3}  &22    &1154
&0.063	&0.000	&0.004	&0.080  &0.059  &0.080  &0.174 &0.059 &-0.002 &0.000 &0.589 &0.511  &0.000  &0.225 &{\bf 0.752} \\ \hline
{\bf Pomeroy-v1}  &34    &859
&0.012	 &-0.032 &-0.020	&-0.006	 &-0.020	&0.008	&-0.026	&-0.020 &0.061 &-0.007	&{\bf 0.589} &-0.014 &0.140  &0.056 &0.067 \\  \hline
{\bf Pomeroy-v2} &42    &1381
&0.502	&0.576	&0.591	&{\bf 0.617}	&0.568	&0.577	&0.602	&0.568	&0.586 &0.376
&0.513 &0.544	&0.582  &0.564 &0.492 \\  \hline
{\bf Ramaswamy} &190    &1365
&0.618	&0.401	 &0.623	&0.651 &0.618	&0.620	&0.663	&0.639	&0.592 &0.336
&0.000 &0.547	&{\bf 0.676}  &0.182 &0.450 \\  \hline
{\bf Risinger} &42    &1773
&0.210	&0.162	&0.297	&0.223  &0.201 &0.258  &0.153 &0.201 &0.309 &0.000 &0.399 &0.308 &0.248 &0.087 &{\bf 0.428}\\  \hline
{\bf Shipp-v1}    &77    &800
&{\bf 0.264}	&0.035	&0.208	&0.132  &-0.002 &0.168 &0.203 &-0.002 &0.113 &0.079 &0.101  &0.069 &0.124  &0.013 &0.065 \\ \hline
{\bf Singh}  &102    &341 &0.048	&0.037	&0.019	&0.033  &-0.003 &0.069 &-0.003 &0.066 &0.079 &0.066 &0.000  &0.029  &0.034  &0.083 &{\bf 0.159}\\  \hline
{\bf Su}   &174    &1573
&0.666	&0.672	&0.662	&0.738  &0.687  &0.650 &0.667 &0.660 &0.657 &0.589 &0.000 &{\bf 0.824} &0.702  &0.288 &0.738 \\  \hline
{\bf Tomlins-v1}  &104   &2317
&0.396	&0.382	&0.454	&0.409	&{\bf 0.647} &0.469 &0.419	&0.590  &0.374  &0.423 &0.000	&0.485	&0.513  &0.165 &0.413\\  \hline
{\bf Tomlins-v2} &92    &1290
&0.368	&0.222	&0.215	&0.292 &0.226	&0.383	&0.354	&0.311	&0.340 &0.000 &0.000 &0.468	&0.373   &{\bf 0.470}  &0.288\\  \hline
{\bf West}  &49    &1200
&0.489	&0.403	&0.489	&0.442		&{\bf 0.515}	&0.489	&0.442	&{\bf 0.515} &0.258 &0.00
&0.459  &0.412	&0.391 &0.016 &0.322 \\  \hline
{\bf Yeohv2}  &248    &2528
&0.385	&0.002	&0.383	&0.479	&0.530	&{\bf 0.550}	&0.337	&0.442	&0.405 &0.000 &0.057  &0.172	&0.465   &-0.001 &0.135 \\  \hline
\hline 
\end{tabular}%
}
\end{table*}


\end{document}